\documentclass[conference,compsoc]{IEEEtran}
\usepackage{cite}
\usepackage{times}
\usepackage{xcolor}
\usepackage{soul}
\usepackage{graphicx}
\usepackage[utf8]{inputenc}

\usepackage{algorithm}
\usepackage{algorithmic}
\usepackage{amsmath}
\usepackage{amsfonts}
\usepackage{graphics}
\usepackage{epsfig}
\usepackage{subfigure}
\usepackage{float}
\usepackage{multirow}

\usepackage{color}
\usepackage{times}
\usepackage{helvet}
\usepackage{courier}
\usepackage{multirow}
\usepackage{bm}
\usepackage{url}
\usepackage{hyperref}

\newtheorem{remark}{Remark}[section]

\def\lf{\left\lfloor}   
\def\rf{\right\rfloor}
\def\etal{{\em et al.\/}}

\begin{document}
\title{Power Law in Sparsified Deep Neural Networks}

\author{Lu Hou \hspace{.1in} James T. Kwok\\
	Department of Computer Science and Engineering\\
	Hong Kong University of Science and Technology\\
	Hong Kong\\
}
\maketitle

% As a general rule, do not put math, special symbols or citations
% in the abstract
\begin{abstract}
	The power law  
	has been observed in the degree distributions of many biological neural networks. 
	Sparse deep neural networks, which learn an economical 
	representation 
	from the data,
	resemble biological neural networks in many ways.  In this paper,  we  study if 
	these artificial networks
	also exhibit properties of the power law. Experimental results on two popular deep learning
	models, namely, multilayer perceptrons and convolutional neural networks,
	are affirmative.
	The power law   is also naturally related to 
	preferential attachment. To study the dynamical properties of deep networks
	in continual learning,
	we propose an internal preferential attachment model to explain how the network topology
	evolves. Experimental results show that
	with the arrival of a new task,
	the new connections made follow this preferential attachment process. 
	%These properties reflect important connectivity topology and formation process of deep neural networks and may shed light to more efficient and economical representation learning in deep neural networks.
\end{abstract}

%%%%%%%%%%%%%%%%%%%%%%%%%%%%%%%%%%%%%%%%%%%%%%%%%%%%%%%%%%%%%%%%%%%%%%

\section{Introduction}

The power law distribution has been commonly used to 
describe 
%and understand 
the underlying mechanisms 
of a wide variety of physical, biological and man-made networks
 \cite{barabasi1999emergence}.
Its probability density function 
%(PDF) 
is of the form:
\begin{equation} \label{eq:law}
f(x) \propto x^{-\alpha}, 
\end{equation} 
where $x$ is the measurement, and $\alpha > 1$ is the exponent.
%\cite{barabasi2003scale}. 
It is well-known that the power law 
can originate in an evolving network via preferential attachment 
\cite{barabasi1999emergence}, in which
new connections are 
preferentially made to the more highly connected nodes.
%in the network.
Networks exhibiting a power law degree distribution are also called scale-free.

%Power law, which describes a functional relationship between two quantities where one quantity varies as a power of another, has been observed in the degree distributions of a wide range of and man-made networks~\cite{barabasi1999emergence,eguiluz2005scale,johannesson2006afterglow}.

In the context of biological neural networks,
the power law and its variants have been commonly observed.
For example, 
Monteiro \etal
\cite{monteiro2016model}
%[\citeyear{monteiro2016model}] 
showed that the mean
learning curves of scale-free networks 
resemble that of the biological neural network of the worm 
{\em Caenorhabditis Elegans}. Moreover, these 
learning curves
are better 
than those generated from random and small-world networks.
Eguiluz \etal
\cite{eguiluz2005scale} 
%[\citeyear{eguiluz2005scale}] 
studied the functional networks connecting
correlated human brain sites, and 
showed  that the distribution of functional connections also
follows the power law.

In practice, few empirical phenomena obey the power law exactly 
\cite{clauset2009power}.  Often, 
the degree distribution has a power law regime followed by  a fall-off. 
This may result from the finite size of the data, temporal limitations of the collected data or constraints imposed
by the underlying physics~\cite{burroughs2001upper}.
Such networks are sometimes called broad-scale or truncated scale-free networks
\cite{amaral2000classes}, and have been observed for example in 
the human brain anatomical network
\cite{iturria2008studying}.
To model this 
upper truncation effect,  extensions of the power law have been proposed
\cite{burroughs2001upper,kolyukhin2013power}. In this paper, we will focus on 
the truncated power law (TPL) distribution proposed in 
\cite{kolyukhin2013power}, which explicitly includes
a lower 
and
upper threshold.

%These cases can be described using the truncated power law distribution instead, in which the connectivity distribution has a power law regime followed by a fall-off.

Recently, deep neural networks have achieved state-of-the-art performance in
various tasks such as speech recognition, visual object recognition, and image
classification~\cite{lecun2015deep}. 
However, connectivity of the network, and subsequently its degree distribution, are fixed by
design. Moreover, 
%as expected, 
many of its 
connections are redundant. 
Recent studies show that these deep networks can often be significantly sparsified.
In particular,
by pruning unimportant connections and then retraining the remaining connections,
the resultant sparse network
often suffers no performance degradation \cite{han2015deep,li2017pruning}. 
This sparsification process is also
analogous to
how learning works 
in the mammalian brain \cite{han2015learning}. For example,
the pruning (resp. retraining) in artificial neural networks
resembles
the weakening (resp. strengthening) of functional connections in brain maturity.

Another similarity between biological and deep neural networks can be seen
in the context of
continual learning,  in  which the network learns progressively with the arrival of new
tasks
\cite{kirkpatrick2017overcoming}.
Biological neural networks are able to learn continually as they evolve over a lifetime
\cite{anderson2010neural}.
Continual learning 
by deep networks 
mimics this biological learning process in which new connections are made without
loss of established functionalities in neural circuits
\cite{fernando2017pathnet}.
%resembles the biological learning process are 
%In PathNet~\cite{fernando2017pathnet},a tournament selection genetic algorithm is used to form new connections for new tasks while fixing those learned for the previous tasks, allowing parameter reuse without catastrophic forgetting. 
In both biological and artificial neural networks, sparsity 
works as a regularizer and
allows a more economical representation of the learning experience
to be obtained.
%According to a neural reuse theory, neural circuits established can be exapted to new tasks without loss of original functionality. 

In general,
a network has both static and dynamical properties \cite{barabasi2002evolution}.
Static properties describe the topology of the network, while dynamical
properties describe the dynamics governing network evolution and
explain how the topology is formed.
In this paper, we study if the 
sparsified deep neural networks also exhibit properties of the power law
as observed in their biological counterparts. 
%For scale-free networks, the static topological property that the degree distribution   follows a power law can be explained by the dynamical preferential attachment process.

The rest of this paper is organized as follows. 
Section~\ref{sec:review} first reviews the power law and preferential attachment.
Section~\ref{sec:powerlaw} studies the static properties 
of sparsified deep neural networks.
In particular, we examine the degree distributions
of two popular deep learning models, namely, multilayer perceptrons and convolutional neural networks,
and show that they follow
the truncated power law.
Section~\ref{sec:pref}
studies the dynamics behind this power law behavior.
We propose a preferential attachment model for deep neural networks,
and verify  that new connections added to the artificial network in a continual learning setting follow
this model.
Finally, the last section gives some concluding remarks.

\section{Related Work}
\label{sec:review}

%%%%%%%%%%%%%%%%%%%%%%%%%%%%%%%%%%%%%%%%

\subsection{Truncated Power Law}

In practice, few empirical phenomena obey the power law 
in (\ref{eq:law})
exactly for the whole range of observations
\cite{clauset2009power}.  Often, 
the power law applies only for values greater than some minimum
\cite{clauset2009power}.
%\cite{kolyukhin2013power}.
There may also be a maximum value above which the power law is no longer valid.
Such an upper truncation is often
observed 
in natural systems 
like forest fire areas, hydrocarbon volumes, 
fault
lengths, and oil and gas field sizes 
\cite{burroughs2001upper}. This
may result from the finite size of the data, 
temporal limitations of the collected data, or constraints imposed
by the underlying physics~\cite{burroughs2001upper}.
For example, the size of forest fires is naturally limited by availability of fuel and climate,
and the upper-bounded power law fits the data better 
\cite{malamud1998forest}.

The truncated power law (TPL) distribution, with lower threshold
$x_{\min}$ and
upper threshold $x_{\max}$, captures the above effects
\cite{kolyukhin2013power}.
Its  
probability density function 
(PDF) and complementary cumulative distribution function
(CCDF)\footnote{The CCDF is defined as one minus the
	cumulative distribution function, i.e.,
	$1 -  \int_{x_{\min}}^{x} f(x)\; dx = \int_{x}^{x_{\max}} f(x)\;dx$.}
are given by:
%\cite{kolyukhin2013power}:
\begin{equation} \label{eq:ccdf}
p(x) = \frac{1-\alpha}{x_{\max}^{1-\alpha} - x_{\min}^{1-\alpha}} x^{-\alpha}, 
\quad 
S(x) = \frac{x^{1-\alpha} - x_{\max}^{1-\alpha}}{x_{\min}^{1-\alpha} - x_{\max}^{1-\alpha}}.
\end{equation}

It is well-known that the log-log CCDF plot for the standard power law distribution is a line.
However,
from (\ref{eq:ccdf}),
the log-log CCDF for a TPL is
\begin{equation} \label{eq:tpl_ccdf} 
\log(S(x)) = \log(x^{1-\alpha} - x_{\max}^{1-\alpha}) - \log(x_{\min}^{1-\alpha} - x_{\max}^{1-\alpha}). 
\end{equation}
As $\lim_{x \to x_{\max}} \log(S(x)) = -\infty$, the log-log CCDF plot for TPL has a fall-off near $x_{\max}$.
Moreover, when $x_{\max}$ is large, $\log(S(x)) \simeq (1-\alpha)\log(x)  - (1-\alpha)\log(x_{\min})$,
and 
the log-log CCDF plot reduces to a line. 
When $x_{\max}$ gets smaller, the linear region shrinks and
the fall-off starts earlier.

When $x$ only takes integer values (instead of a range of continuous values),
the probability function of TPL distribution becomes~\cite{hanel2017fitting}
\begin{equation} \label{eq:discrete}
f(x) = \frac{ x^{-\alpha}}{\zeta(\alpha, x_{\min}, x_{\max})}, 
\end{equation}
where $\zeta(\alpha, x_{\min}, x_{\max}) = \sum_{k=x_{\min}}^{x_{\max}} k^{-\alpha}$.  The corresponding CCDF is:
	$S(x) = \frac{\zeta(\alpha, x, x_{\max})}{\zeta(\alpha, x_{\min}, x_{\max})}$.

%%%%%%%%%%%%%%%%%%%%%%%%%%%%%%%%%%%%%%%%

\subsection{Preferential attachment}
\label{sec:prefer_review}

The power law distributions
can originate from the process of preferential attachment 
\cite{barabasi1999emergence},
which can be
either 
external 
or 
internal 
\cite{barabasi2002evolution}.
External preferential attachment refers to that when a new node is added to the network, it is more
likely to connect to an existing node with high degree;
while internal preferential attachment means that existing nodes with high degrees are more likely
to connect to each other.  
In this paper  (as 
will be explained in Section~\ref{sec:pref}), 
we will focus
on 
internal preferential attachment.

Let $N$ be the number of nodes in the network, and $a$ be the number of new
internal connections 
created 
in unit time
per existing node.
For two nodes with degrees $d_1$ and $d_2$, 
the expected number of new connections created between  them per unit time is:
\begin{equation}
\label{eq:internal}
\Delta(d_1, d_2) = 2Na \frac{d_1d_2}{\sum_{s,m\neq s} d_sd_m},
\end{equation}
where $s$ and $m$ are the indices to all the nodes in the network.

%%%%%%%%%%%%%%%%%%%%%%%%%%%%%%%%%%%%%%%%%%%%%%%%%%%%%%%%%%%%%%%%%%%%%%

\section{Power Law in Neural Networks}
\label{sec:powerlaw}

In this section, we study whether the degree distributions in artificial neural networks
follow the power law.
However, 
connectivity of the
network is often fixed by design and not learned from data. 
%investigation of the power law behavior is thus inconsequential. 
Moreover, many of the network connections are redundant. Hence, we will
study networks that have been sparsified, in which only important connections are kept.
Specifically, 
we use sparse networks produced by the state-of-the-art
three-step network pruning method
in~\cite{han2015learning},
and also pre-trained
sparse  convolutional neural networks.
%(AlexNet and VGG-16).
For the three-step network pruning method,
a dense network
is first trained.
For each layer,
a fraction of $s
\in (0,1)$ 
connections with the smallest magnitudes
are pruned.
The unpruned connections
are then retrained.
To avoid potential performance degradation, 
connections to the output layer is always left unpruned, as is common in network pruning~\cite{li2017pruning} and
continual learning~\cite{fernando2017pathnet}.

Obviously, neural networks are of finite size and the connections each node can make
is limited. 
This suggests upper truncation and the TPL is more appropriate for modeling connectivity.
In particular, 
the discrete TPL is more suitable
as the degree is integer-valued.
Moreover,
as the number of nodes, the  maximum number of connections each node can make, and the
nature of features extracted at different layers
are different, 
it is more appropriate to study the connectivity in a layer-wise manner.

%To fit the TPL,
%we assume that the central $(100-2k)\%$ of the degree values 
%in each layer
%follow the power law. In other words,
%let $n$ be the number of nodes in that layer.
%the smallest ($\hat{x}_{\min}$)
%and largest 
%($\hat{x}_{\max}$)
%degree values for fitting are
%the $\lf(n \times k\%)\rf$th smallest, and
%$\lf(n \times k\%)\rf$th largest degree values, respectively.
%We then regress $\log(S(x))$ 
%in (\ref{eq:tpl_ccdf})
%against $\log(x)$ as in \cite{burroughs2001upper},
%with
%$x_{\min} =
%\hat{x}_{\min}$ and $x_{\max}=\hat{x}_{\max}+1$. We add one to 
%$x_{\max}$ so that
%$\log(S(x))$ will not go to infinity when it approaches $x_{\max}$. Empirically, $k$ is chosen
%from $\{5, 7.5, 10\}$,
%and the one which
%minimizes the mean squared error is used.

We follow the method in \cite{clauset2009power} 
to estimate $x_{\min}$, $x_{\max}$ and $\alpha$ in (\ref{eq:discrete}).
Specifically, $x_{\min}$ and $x_{\max}$ are chosen by minimizing the difference
between the probability distribution of the observed data and the best-fit
power-law model as measured by the Kolmogorov-Smirnov (KS) statistic:
%\[ D = \text{max}_{x_{\min} \leq x \leq x_{\max}} |S(x) - P(x)|, \]
\[ D = \text{max}_{x\in \{x_{\min}, 
	x_{\min} +1, \dots, x_{\max}-1,x_{\max}\}} |S(x) - P(x)|, \]
where $S(x)$ and $P(x)$ are the CCDF's of the observed data and fitted power-law
model 
for $x\in \{x_{\min} ,
	x_{\min} +1, \dots, x_{\max}-1,x_{\max}\}$.
%in the range $x_{\min} \leq x \leq x_{\max}$. 
To reduce the search space, 
we search
$x_{\min}$ in the $\lf(n \times k\%)\rf$ smallest degree values, 
where $n$ is the number of nodes in that layer,
and 
$x_{\max}$ in the 
$\lf(n \times k\%)\rf$ largest degree values, respectively.
%Empirically, $(k_1,k_2)=(20,5)$ is used for MLP  and $(k_1, k_2) = (30, 30)$ for CNN.
Empirically,  $k=30$ is used.
For each 
$(x_{\min}, x_{\max})$
pair, we estimate $\alpha$ using the method of maximum likelihood as in ~\cite{clauset2009power}.
%with $x_{\min} = \hat{x}_{\min}$ and $x_{\max}=\hat{x}_{\max}$.
%The pair that  minimizes the KS statistic is chosen as the final estimation of $\hat{x}_{\min}$ and $\hat{x}_{\max}$.
%Following~\cite{kolyukhin2013power}, 
%we require $\alpha>1$.

%To evaluate the goodness of fit, we compute the p-value 
%by Monte Carlo simulations 
%\cite{deluca2013fitting}.
%Specifically, a number of 
%synthetic sample power-law distributions with the same $(x_{\min},
%x_{\max}, \alpha)$ as the fitted model 
%are generated
%using the inversion 
%%or transformation 
%method~\cite{deluca2013fitting}:
%\[
%x_i = \left\lfloor \frac{x_{\min}}{(1 - (1-r^{\alpha-1}) u_i)^{1/{\alpha-1}}}
%\right\rfloor,
%\]
%where $r= \frac{x_{\min}}{x_{\max}}$, 
% $u_i$ is randomly 
%sampled
%from 
%the uniform distribution in
%$[0,1)$,
%and $\lfloor \cdot \rfloor$ denotes the floor operator.
%%operator is used to generate only integer degrees\footnote{The original method is for sampling continuous data.}.
%%The simulation is run many times and 
%The $p$-value is defined as the fraction of 
%KS distances
%obtained from the synthetic data
%that are larger than the empirical KS distance 
%obtained from  the observed data.
%%Following ~\cite{clauset2009power}, 
%We accept the hypothesis  that the data is from the estimated TPL 
%if $p > 0.05$,
%and reject 
%otherwise.

%%%%%%%%%%%%%%%%%%%%%%%%%%%%%%%%%%%%%%%%

\subsection{Multilayer Perceptron (MLP) on MNIST}
\label{sec:mnist-fc}

In this section, we perform experiment on the MNIST data set\footnote{\url{http://yann.lecun.com/exdb/mnist/}},
which contains $28 \times 28$ gray images from 10 digit classes.  $50,000$ images are used for
training, another $10,000$ for validation, and the remaining $10,000$  for testing.

Following \cite{han2015learning}, we 
first train a dense 
MLP
(with two hidden layers)
which uses the cross-entropy loss
in the Lasagne package\footnote{\url{https://github.com/Lasagne/Lasagne/blob/master/examples/mnist.py}}:
\[ \text{\em input} - 1024\text{\em FC} - 1024\text{\em FC}- 10\text{\em softmax}. \]
Here, 
$1024\text{\em FC}$ denotes a fully connected layer with 1024 units
and
$10\text{\em softmax}$ is a softmax layer 
for the 10 classes.
The optimizer is
stochastic gradient descent (SGD) with momentum.
All hyperparameters are the same as in
the Lasagne package.
The maximum number of epochs is $200$.
After training, a fraction of $s =0.9$ connections are pruned in each layer.
The testing accuracies of the dense and (retrained) pruned networks are comparable
($98.09\%$ and    $98.21\%$).

\begin{figure}[htbp]
	\begin{center}
		%({\sf K}=400). 
		\subfigure[input.
		%({\sf K}=1024). 
		\label{fig:mnist_1024_1}]{\includegraphics[width=0.155\textwidth]{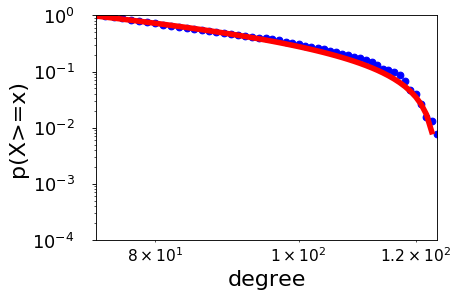}}
		\subfigure[fc1.
		%({\sf K}=1024).
		\label{fig:mnist_1024_2}]{\includegraphics[width=0.155\textwidth]{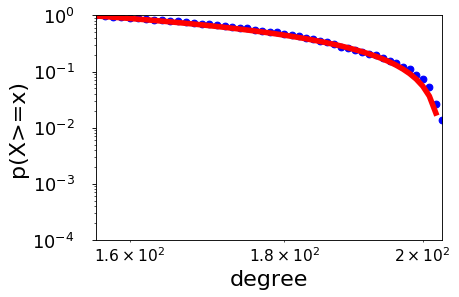}}
		\subfigure[fc2.
		%({\sf K}=1024).
		\label{fig:mnist_1024_3}]{\includegraphics[width=0.155\textwidth]{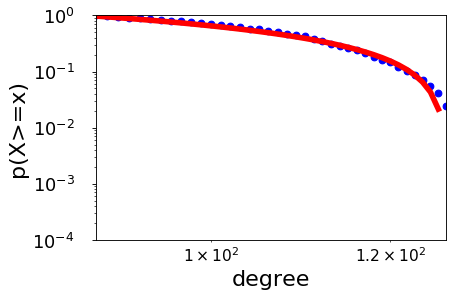}}		
		%({\sf K}=2048).
\caption{ Log-log CCDF plot and TPL fit (red) for the MLP layers on MNIST. 
			``fc1" and
			``fc2" 
			denote the first and 
			second FC layer, respectively.  The 
			softmax layer,
			which is not pruned,
			is not shown.}
		\label{fig:fc_ccdf}
	\end{center}
\end{figure}

For each node, we obtain its
degree 
by counting the total number of connections (after pruning) to nodes in its upper and lower
layers.\footnote{For a node in the input (resp. last pruned) layer, we only count its
	connections to nodes in the upper (resp. lower) layer.}
Figure~\ref{fig:fc_ccdf} 
%and Table~\ref{tbl:mnist_mlp_fitting} 
shows
the CCDF plot
and TPL fit for each layer in the MLP.
%, respectively.
As can be seen, the TPL fits the degree distributions well.

%\begin{table}[htbp]
%	\centering
%\caption{Power-law fits for different MLP layers (on MNIST
%data) and their $p$-values.}
%	\label{tbl:mnist_mlp_fitting}
%	\begin{tabular}{cc|ccc|c}
%		\hline
%		layer & $n$  & $x_{\min}$ & $x_{\max}$ & $\alpha$ & $p$  \\ \hline
%		input & 784  &        73        &       124        &  2.994   & 0.84 \\
%		 fc1  & 1024 &       156        &       203        &  1.054   & 0.73 \\
%		 fc2  & 1024 &        89        &       127        &  1.062   & 0.40 \\ \hline
%	\end{tabular}
%\end{table}

%%%%%%%%%%%%%%%%%%%%%%%%%%%%%%%%%%%%%%%%

\subsection{Convolutional Neural Network on MNIST}
\label{sec:mnist-cnn}

In this section, we use the
convolutional neural network (CNN) 
in the Lasagne package.
It 
is similar to LeNet-5 \cite{lecun1998gradient}, and
has two convolutional layers  followed by 2 fully connected layers:
\[
\text{\em input} - 32C5 - \text{\em MP}2 - 32C5 - \text{\em MP}2 -
256\text{\em FC}- 10\text{\em softmax}
\]
Here, $32C5$ denotes a ReLU convolution layer with $32$ $5 \times 5$ filters, and $\text{\em
	MP}2$ is a $2 \times 2$ max-pooling layer. 
We use SGD with momentum as the optimizer.
The other hyperparameters are the same as in the Lasagne package.
The maximum number of epochs is $200$.
The (dense) CNN has
a testing accuracy of 
98.85\%.
This is then pruned 
using the method in \cite{han2015learning}, 
with 
$s=0.7$.
After retraining,
the sparse CNN 
has a testing accuracy of 
98.78\%, and is comparable with the dense network.

For a convolutional layer,
we consider each feature map at the convolutional layer as a node. 
As in Section~\ref{sec:mnist-fc}, the node degree  is obtained by counting the
total number of connections (after pruning) to its upper and lower layers. As an example, 
consider  
a feature map (node) in the first convolutional layer (conv1).
As the input image is of size $28 \times 28$, each such feature map 
is of size $24\times 24$.
To illustrate  the counting more easily, we consider
the 
unpruned
network.
We first count its connections to the input layer. 
Recall that in conv1,
(i) the filter size 
is $5 \times 5$; 
(ii) each filter weight is used $24 \times
24 = 576$ times; and 
(iii) there is only one channel in the  grayscale MNIST image.  
Thus, each node has
$25 \times 576 \times 1 = 14,400$ connections to the input layer. 
Similarly, for connections to the conv2 layer, 
(i) the conv2 filter size is $5 \times 5$; 
(ii) each filter weight is used $8 \times 8 = 64$ times (size of each conv2 feature map); 
and (iii) there are 32 feature maps in conv2.
Thus, each conv1 node has $25 \times 64 \times 32= 51,200$ connections 
to the conv2 layer.
Hence, the degree of each conv1 node (in an unpruned network) is $14,400+51,200=65,600$.
Note that the pooling layers do not have learnable connections. They are never
sparsified, and we do not need to study their degree distributions.

\begin{figure}[htbp]
	\begin{center}
		\subfigure[conv1. \label{fig:mnist_cnn_conv1}]{\includegraphics[width=0.155\textwidth]{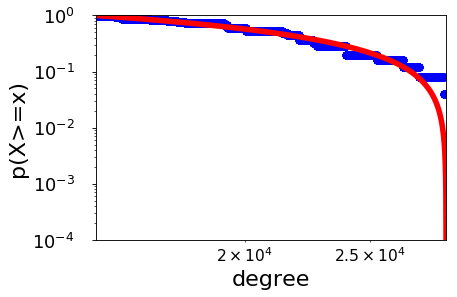}}
		\subfigure[conv2.\label{fig:mnist_cnn_conv2}]{\includegraphics[width=0.155\textwidth]{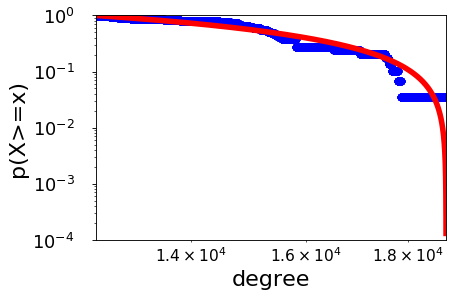}}
		\subfigure[fc3.\label{fig:mnist_cnn_fc1}]{\includegraphics[width=0.155\textwidth]{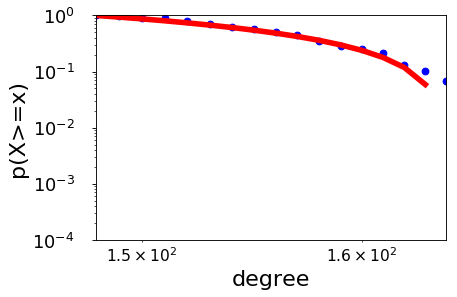}}
		\caption{ Log-log CCDF plot and TPL fit (red) for the CNN layers on MNIST.  Here, ``conv1" and
			``conv2" denote the first and 
			second convolutional layers, and ``fc3" is the 3rd 
			(FC)
			layer.
			%The blue curve represents the CCDF plot $(P(X \geq x))$ while the red curve represents the corresponding TPL fit in range $[x_{min}, x_{max}]$. 
		}
		\label{fig:mnist_ccdf}
	\end{center}
\end{figure}

Figure~\ref{fig:mnist_ccdf} 
%and Table~\ref{tbl:mnist_cnn_fitting} 
shows
the CCDF plot 
and TPL fit for each 
CNN layer.
%, respectively.
Again, the TPL fits the distributions well.

%\begin{table}[htbp]
%	\centering
%	\caption{Power-law fits for different CNN layers (on MNIST data) and their $p$-values.}
%	\label{tbl:mnist_cnn_fitting}
%	\begin{tabular}{cc|ccc|c}
%		\hline
%		layer & $n$ & $x_{\min}$ & $x_{\max}$ & $\alpha$ & $p$  \\ \hline
%		conv1 & 32  &   15360    &   28608    &     1.049      & 0.87 \\
%		conv2 & 32  &   12544    &   18816    &     1.030      & 0.72 \\
%		 fc3  & 256 &    148     &    164     &     1.244      & 0.32 \\ \hline
%	\end{tabular}
%\end{table}

%%%%%%%%%%%%%%%%%%%%%%%%%%%%%%%%%%%%%%%%

\subsection{CNN on CIFAR-10}
\label{sec:cifar-cnn}

\begin{figure}[htbp]
	\begin{center}
		\subfigure[conv1. \label{fig:cifar_cnn_conv1}]{\includegraphics[width=0.155\textwidth]{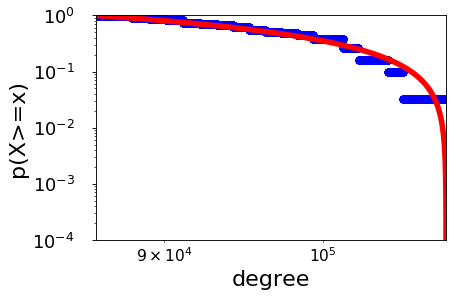}}
		\subfigure[conv2.\label{fig:cifar_cnn_conv2}]{\includegraphics[width=0.155\textwidth]{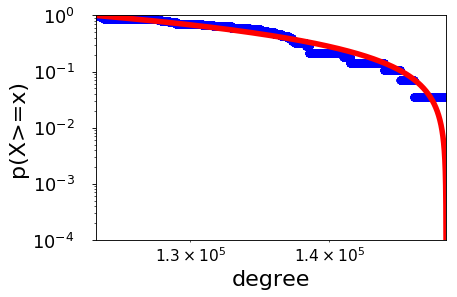}}
		\subfigure[conv3.\label{fig:cifar_cnn_conv3}]{\includegraphics[width=0.155\textwidth]{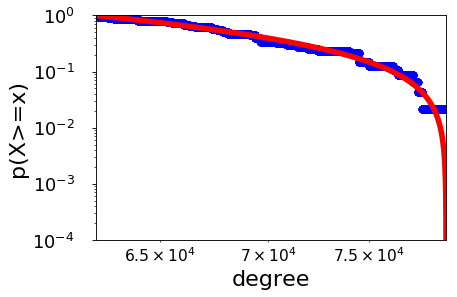}}
		\subfigure[conv4.\label{fig:cifar_cnn_conv4}]{\includegraphics[width=0.155\textwidth]{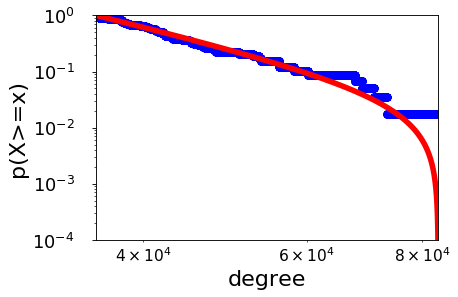}}
		\subfigure[fc5.\label{fig:cifar_cnn_fc1}]{\includegraphics[width=0.155\textwidth]{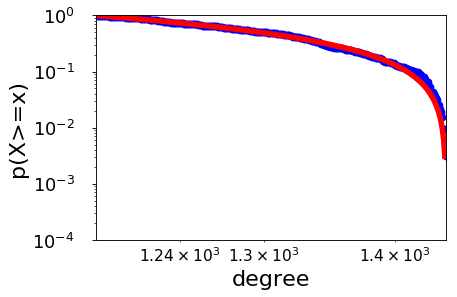}}		
		\caption{ Log-log CCDF plot and TPL fit (red) for the CNN layers on CIFAR-10.}
		\label{fig:cifar10_ccdf}		
	\end{center}
\end{figure}

In this section, we perform experiments on the CIFAR-10 data set\footnote{\url{https://www.cs.toronto.edu/~kriz/cifar.html}},
which contains 
$32 \times 32$ color images from ten object classes.
We use $45,000$ images for training, another $5,000$ for validation, and the remaining $10,000$ for testing. 
The following CNN 
from \cite{zenke2017continual}
is used\footnote{\url{https://github.com/fchollet/keras/blob/master/examples/cifar10\_cnn.py}}:
\begin{eqnarray*}
	&& \text{\em input} - 32C3 - 32C3 -  \text{\em MP}2 - 64C3 - 64C3 - 
	\text{\em MP}2 
	\\ && 
	- 512\text{\em FC}- 10\text{\em softmax}.
\end{eqnarray*}
We use RMSprop as the optimizer,
and the maximum number of epochs is $100$.
The (dense) CNN has 
a testing accuracy of 
81.90\%.
This is then pruned 
using the method in~\cite{han2015learning}, 
with 
$s=0.7$.
After retraining, the testing accuracy of the sparse CNN is 80.46\%, and is comparable with the original network.
Figure~\ref{fig:cifar10_ccdf}
%and Table~\ref{tbl:cifar10_fitting} 
shows
 the CCDF plot
and TPL fit for each
layer.
%, respectively. 
Again, the TPL fits the distributions well.

%
%\begin{table}[htbp]
%	\centering
%\caption{Power-law fits for different CNN layers (on CIFAR-10 data) and their $p$-values.}
%	\label{tbl:cifar10_fitting}
%	\begin{tabular}{cc|ccc|c}
%		\hline
%		layer & $n$ & $x_{\min}$ & $x_{\max}$ & $\alpha$ & $p$  \\ \hline
%		conv1 & 32  &      86016       &      108544      &  1.001   & 0.77 \\
%		conv2 & 32  &      123648      &      148992      &  3.719   & 0.47 \\
%		conv3 & 64  &      62208       &      79104       &  4.905   & 0.77 \\
%		conv4 & 64  &      35584       &      83200       &  4.998   & 0.73 \\
%		 fc5  & 512 &       1183       &       1441       &  2.214   & 0.68 \\ \hline
%	\end{tabular}
%\end{table}

%%%%%%%%%%%%%%%%%%%%%%%%%%%%%%%%%%%%%%%%

\subsection{AlexNet on ImageNet} 

\begin{figure*}[htbp]
	\begin{center}
		\subfigure[conv1. \label{fig:alexnet_conv1}]{\includegraphics[width=0.11\textwidth]{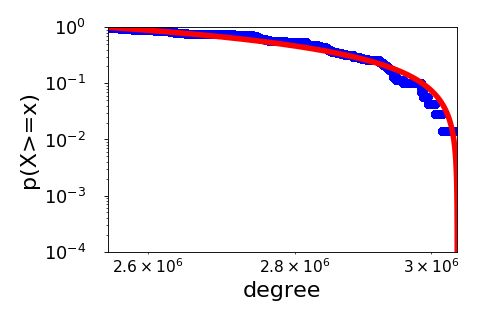}}
		\subfigure[conv2.\label{fig:alexnet_conv2}]{\includegraphics[width=0.11\textwidth]{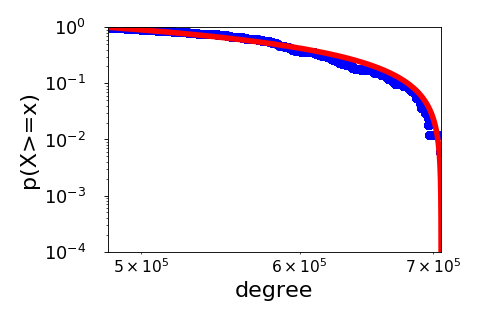}}
		\subfigure[conv3.\label{fig:alexnet_conv3}]{\includegraphics[width=0.11\textwidth]{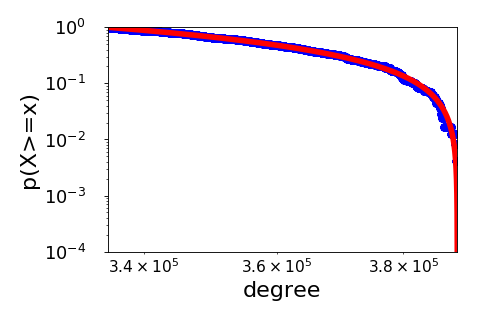}}
		\subfigure[conv4.\label{fig:alexnet_conv4}]{\includegraphics[width=0.11\textwidth]{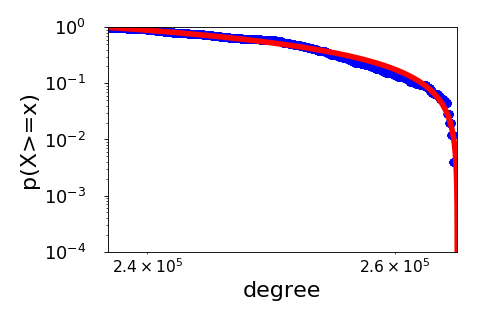}}
		\subfigure[conv5.\label{fig:alexnet_conv5}]{\includegraphics[width=0.11\textwidth]{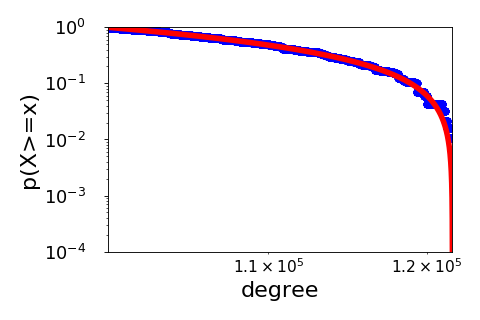}}
		\subfigure[fc6.\label{fig:alexnet_fc6}]{\includegraphics[width=0.11\textwidth]{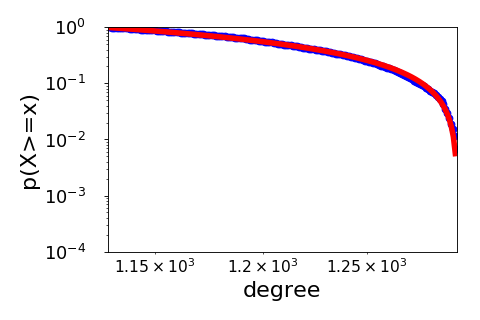}}
		\subfigure[fc7.\label{fig:alexnet_fc7}]{\includegraphics[width=0.11\textwidth]{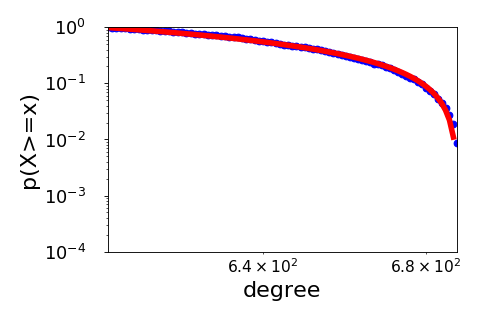}}
		\subfigure[fc8.\label{fig:alexnet_fc8}]{\includegraphics[width=0.11\textwidth]{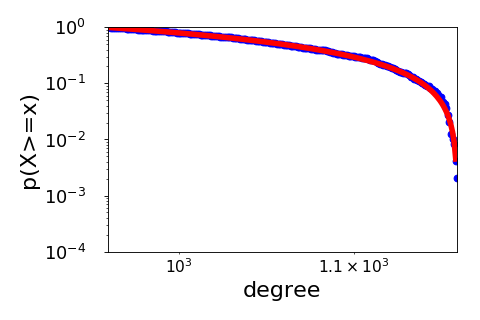}}		
		\caption{Log-log CCDF plot  and TPL fit (red) for the sparse AlexNet layers. 
			Naming of the layers follows that defined in Caffe.
		}
		%The fitted exponent is shown at the title of each subplot.}
		\label{fig:alexnet_ccdf}	
	\end{center}
\end{figure*}

In this section, we perform experiments on 
the ImageNet data set~\cite{russakovsky2015imagenet}, which has
1,000 categories, over 1.2 million 
training images, 50,000 
validation images, and 100,000 test images. 
We use the 
sparsified  
AlexNet\footnote{\url{https://github.com/songhan/Deep-Compression-AlexNet}}
(with 5 convolutional layers and 3 fully connected layers)
%	Note that the weight of two parts in conv1,
%	conv3, conv4 layer are shared.
%\cite{krizhevsky2012imagenet}
from \cite{han2015deep}.
%This model compresses AlexNet from 233MB to 8.9MB without loss of accuracy. 
Figure~\ref{fig:alexnet_ccdf} 
%and Table~\ref{tbl:AlexNet_fitting} 
shows 
the CCDF plot
and TPL fit for the various AlexNet layers.
%, respectively.
%Again, the TPL fits the node degrees well.

%As can be seen from Table~\ref{tbl:AlexNet_fitting}, the $p$-values of fc7 is relatively low.
%some convolution layers (conv1, conv2)
%and fully-connected layers  (fc6 and fc7) are 
%deep networks with feature hierarchy, 
%The earlier layers are more sensitive to various input while the topology of higher convolutional layers is relatively more robust. 

%\begin{table}[htbp]
%	\centering
%	\caption{Power-law fits for different AlexNet layers (on ImageNet data) and their $p$-values. 
%	Fits with $p \leq 0.05$ are in italic.}
%	\label{tbl:AlexNet_fitting}
%	\begin{tabular}{cc|ccc|c}
%		\hline
%		    layer      &      $n$       &    $x_{\min}$    &    $x_{\max}$    &     $\alpha$     &      $p$       \\ \hline
%		conv1 &  96   & 2547498 & 3039920 & 1.001 & 0.09  \\
%		conv2 & 256   & 480969  & 706305  & 1.001 & 0.09 \\
%		    conv3      &      384       &      334789      &      388869      &      3.017       &      0.99      \\
%		    conv4      &      384       &      236938      &      265330      &      1.134       &      0.11     \\
%		    conv5      &      256       &      100724      &      121680      &      3.251       &      0.99      \\
%		 fc6  & 4096  & 1129 & 1295 & 1.007 & 0.05 \\
%		 \textit{fc7}  & \textit{4096}  & \textit{604}  & \textit{688}  &  \textit{1.001}  & \textit{0.01}  \\
%		     fc8       &      1000      &       962        &       1164       &      1.001       &      0.95      \\ \hline
%	\end{tabular}
%\end{table}

%%%%%%%%%%%%%%%%%%%%%%%%%%%%%%%%%%%%%%%%

\subsection{VGG-16 on ImageNet} 

\begin{figure*}
	\begin{center}
		\subfigure[conv1\_1. \label{fig:vgg16_conv1_1}]{\includegraphics[width=0.11\textwidth]{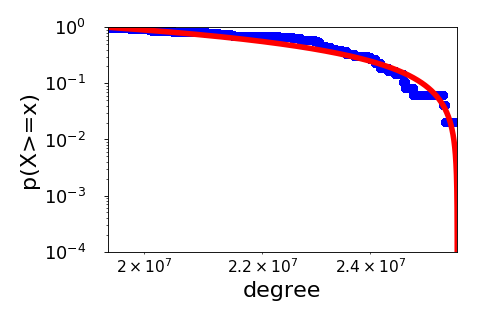}}
		\subfigure[conv1\_2.\label{fig:vgg16_conv1_2}]{\includegraphics[width=0.1100\textwidth]{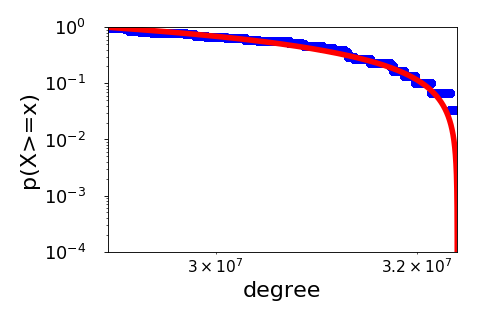}}
		\subfigure[conv2\_1.\label{fig:vgg16_conv2_1}]{\includegraphics[width=0.1100\textwidth]{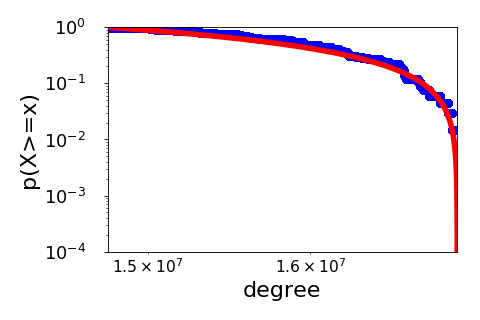}}
		\subfigure[conv2\_2.\label{fig:vgg16_conv2_2}]{\includegraphics[width=0.1100\textwidth]{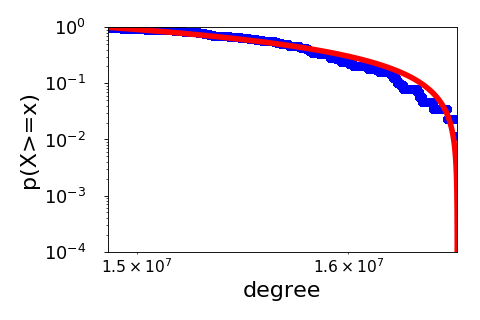}}
		\subfigure[conv3\_1.\label{fig:vgg16_conv3_1}]{\includegraphics[width=0.1100\textwidth]{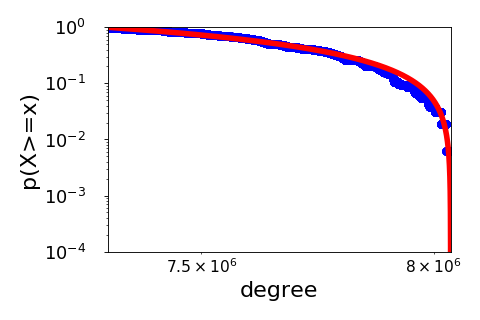}}
		\subfigure[conv3\_2.\label{fig:vgg16_conv3_2}]{\includegraphics[width=0.1100\textwidth]{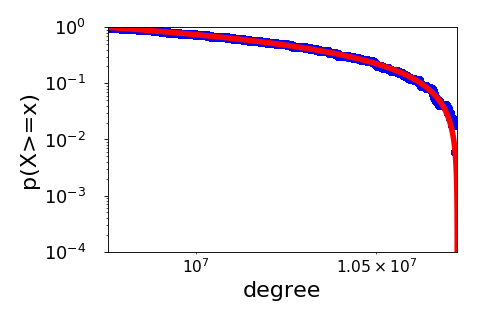}}
		\subfigure[conv3\_3.\label{fig:vgg16_conv3_3}]{\includegraphics[width=0.1100\textwidth]{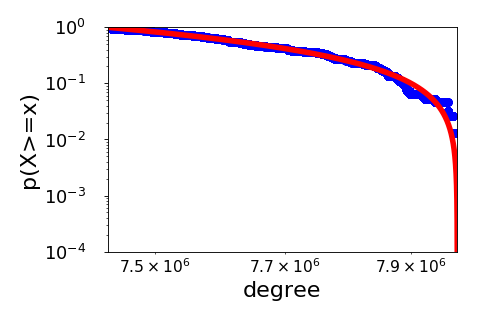}}
		\subfigure[conv4\_1.\label{fig:vgg16_conv4_1}]{\includegraphics[width=0.1100\textwidth]{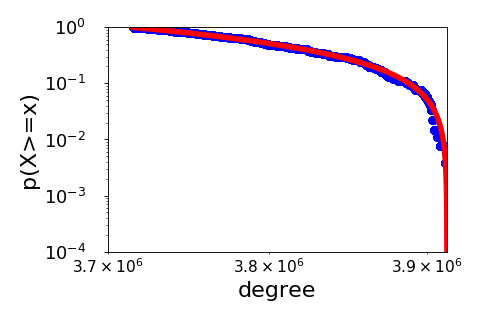}}
		\subfigure[conv4\_2.\label{fig:vgg16_conv4_2}]{\includegraphics[width=0.1100\textwidth]{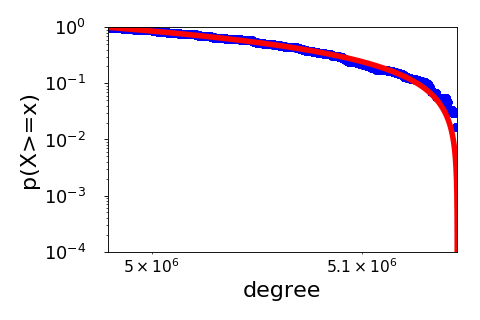}}
		\subfigure[conv4\_3.\label{fig:vgg16_conv4_3}]{\includegraphics[width=0.1100\textwidth]{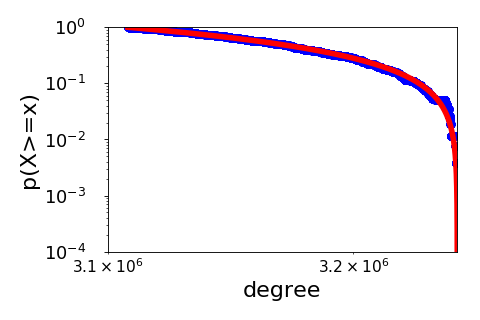}}
		\subfigure[conv5\_1.\label{fig:vgg16_conv5_1}]{\includegraphics[width=0.1100\textwidth]{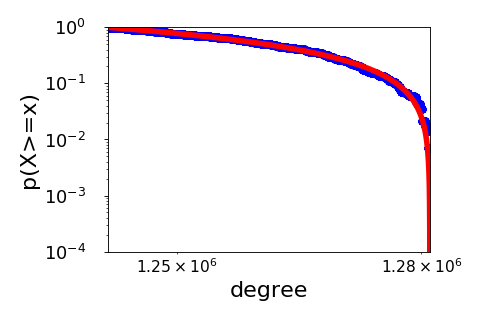}}
		\subfigure[conv5\_2.\label{fig:vgg16_conv5_2}]{\includegraphics[width=0.1100\textwidth]{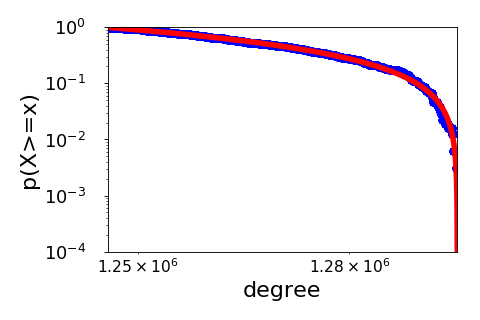}}
		\subfigure[conv5\_3.\label{fig:vgg16_conv5_3}]{\includegraphics[width=0.1100\textwidth]{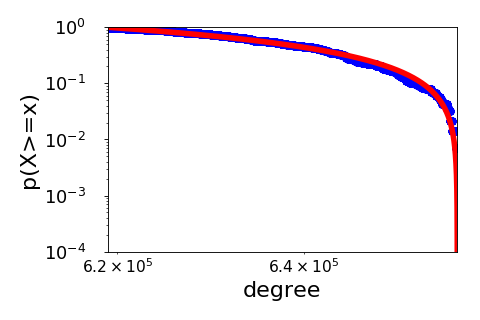}}
		\subfigure[fc6.\label{fig:vgg16_fc6}]{\includegraphics[width=0.1100\textwidth]{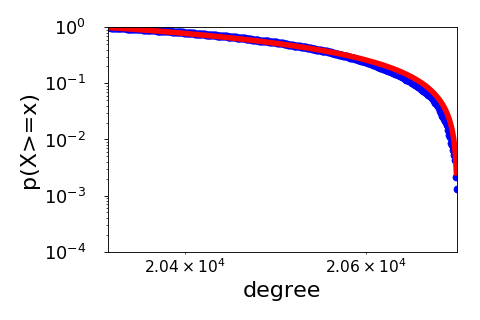}}
		\subfigure[fc7.\label{fig:vgg16_fc7}]{\includegraphics[width=0.1100\textwidth]{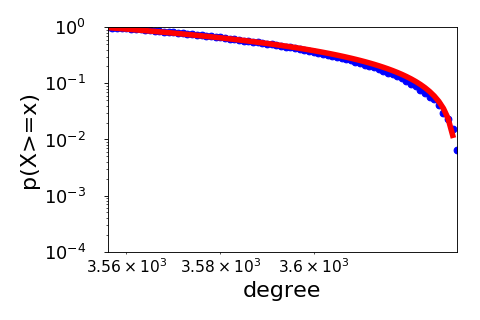}}		
		\subfigure[fc8.\label{fig:vgg16_fc8}]{\includegraphics[width=0.1100\textwidth]{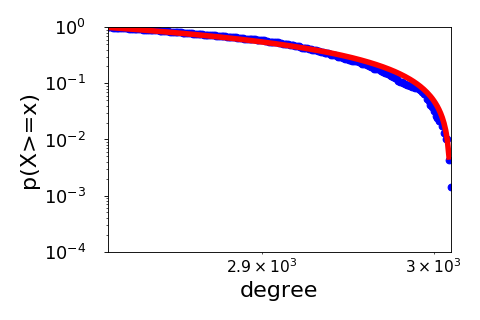}}		
		\caption{Log-log CCDF plot and TPL
			fit (red) for the VGG-16 layers. 
			Naming of the layers follows that defined in Caffe.
		}
		\label{fig:vgg16_ccdf}	
	\end{center}
\end{figure*}

In this section, 
we use
the (dense) VGG-16 
model\footnote{\url{https://github.com/songhan/DSD/tree/master/VGG16}}
(with 13 convolutional layers and 3 fully connected layers)
obtained by the dense-sparse-dense (DSD) procedure of
\cite{han2016dsd}.
We then prune this using the method in
\cite{han2016dsd} with $s=0.3$.
The top-1 and top-5 testing accuracies for the original dense CNN are 68.50\% and 88.68\%, respectively, while those of the sparse CNN are 71.81\% and 90.77\%.
Figure~\ref{fig:vgg16_ccdf} 
%and Table~\ref{tbl:vgg16_fitting} 
shows
the CCDF plot  and
TPL fit for various layers.
%, respectively.
%As can be seen
%from Table~\ref{tbl:AlexNet_fitting}, 
%the TPL fits the distributions well, except
%%the $p$-values of the first  convolutional layer and 
%possibly for the fully-connected layers.\footnote{Note that even when the data
%follows exactly the power-law distribution with the estimated exponent, we will
%still reject the hypothesis in $5\%$ of the cases for a rejection threshold of
%$p\leq 0.05$ \cite{deluca2013fitting}.}
%relatively low.

%{\color{red} except for the first four  convolutional layers.  An obvious gap is observed between the fit and the original CCDF plot for these four plots even with the lower-bound of the exponent $\alpha=1$.  This may because the higher layer topology are more robust in deep networks, as shall be shown and explained in the experiments in Section~\ref{sec:expt}.  }

\section{Internal Preferential Attachment in NN}
\label{sec:pref}

To study the 
dynamical properties of
neural network
connectivity,
we consider 
the continual learning setting 
in which consecutive tasks are learned
\cite{kirkpatrick2017overcoming}.
%without forgetting how to perform previously trained tasks.
Specifically,
at time $t=0$,
an initial sparse network 
is trained
to learn the first task.
At each following timestep $t=1,2,\dots$, 
a new task
is encountered, and the network is re-trained on the new task.
We assume that  
only new connections,
but not new nodes, can be added.
Hence, 
preferential attachment, if exists,
can only be internal 
but not
external.

%%%%%%%%%%%%%%%%%%%%%%%%%%%%%%%%%%%%%%%%

\subsection{Evolution of the Degree Distribution}

In this section, we 
%focus on feedforward networks, and
assume that only nodes in adjacent layers can be connected, as is common in feedforward
neural networks.
Consider a 
pair
of nodes, one from layer $l$ with degree $d_1$ and the other from layer $(l+1)$ with degree $d_2$.
If internal preferential attachment exists,  
the number of connections between this node pair grows proportional to the product $d_1d_2$. 
Analogous to (\ref{eq:internal}), the number of connections created per unit time between
this node pair 
at time $t$ is:\footnote{Unlike (\ref{eq:internal}), note that there is no factor of 2 here.}
\begin{equation} \label{eq:new}
\Delta_t^l(d_1,d_2) = N^la^l \frac{d_1d_2}{\sum_{s,m}d_sd_m}, 
\end{equation} 
where $s$ and $m$ are indices to all the nodes in the two
layers involved,
$N^l$ is the number of nodes in layer $l$, and $a^l$ is the number of new internal connections created in unit time 
from a node in layer $l$ to 
layer
$(l+1)$.

Next, we study how the degree of a node evolves.
For node $i$ at layer $l$,
let $d_i(t)$ be its degree 
at time $t$. Out of these 
$d_i(t)$ connections, let
$d_i^{\uparrow}(t)$ be connected to the upper layer, and $d_i^{\downarrow}(t)$ 
to the lower layer\footnote{For simplicity, we only consider hidden layers
	here.
	Analysis for the other layers can be
	easily modified and are not detailed here.}.
Using (\ref{eq:new}),
the increase of its degree due to new connections to the upper layer is:
\begin{eqnarray*} 
	\frac{\mathrm d d_i^{\uparrow}(t)}{\mathrm d t} &=& \sum_m \Delta^l_t(d_i(t),d_m(t)) \\
	%&=& N^la^l\sum_m \frac{d_i^l(t)d_m^{l+1}(t)}{\sum_{s,m}d_s^l(t)d_m^{l+1}(t)} \nonumber\\
	&=& N^la^ld_i(t) \frac{\sum_m d_m(t)}{(\sum_s d_s(t))(\sum_m d_m(t))} \\
	&=&\frac{ N^la^ld_i(t) }{\sum_s d_s(t)},
\end{eqnarray*}
where $m$ and $s$ are indices to all the nodes in layer $(l+1)$  and layer
$l$, respectively.
Similarly,
the increase of degree due to new connections to the lower layer is:
\[ \frac{\mathrm d d_i^{\downarrow}(t)}{\mathrm d t} =\sum_r \Delta^{l-1}_t(d_r(t), d_i(t)) 
=  \frac{ N^{l-1}a^{l-1}d_i(t) }{\sum_s d_s(t)},\]
where $r$ and $s$ are indices to all the nodes in layer $(l-1)$  and layer
$l$, respectively.
The total number of new connections for nodes in layer $l$ can be obtained as:
\begin{eqnarray*}
	\sum_s d_s(t) &=& \sum_s d_s(0) + \int_{0}^{t} (N^la^l+N^{l-1}a^{l-1}) dt\\
	&=& \sum_s d_s(0) + (N^la^l+N^{l-1}a^{l-1}) t.
\end{eqnarray*}
Combining all these,  we have
\begin{eqnarray}
\frac{\mathrm d d_i(t)}{\mathrm d t} 
&=& \frac{\mathrm d d_i^{\uparrow}(t)}{\mathrm d t}+\frac{\mathrm d  d_i^{\downarrow}(t)}{\mathrm d t} \nonumber\\
&=&  \frac{ (N^la^l+N^{l-1}a^{l-1})  d_i(t)}{\sum_s d_s(0) + (N^la^l+N^{l-1}a^{l-1}) t} \label{eq:tmp}.
\end{eqnarray}
After integration and simplification, we obtain
\begin{equation} \label{eq:dist-d}
d_i(t) = d_i(0) c^l(t),
\end{equation} 
where 
$c^l(t)=1+\frac{(N^la^l+N^{l-1}a^{l-1}) t}{\sum_s d_s^l(0)}$.
Hence, $d_i(t)$ is linear w.r.t. the node's initial degree $d_i(0)$.
In particular, assume that
the degree distribution 
of layer $l$
at $t=0$
(denoted $p_0^l$)
follows the power law  (standard or TPL),
i.e.,
$p_0^l(d) = A d^{-\alpha}$
for some $A > 0$ and $\alpha > 1$. Then,
from
(\ref{eq:dist-d}),
its degree distribution 
at time $t$ is:
\begin{equation} \label{eq:degree-t}
p_t^l(d) = p_0^l \left(\frac{d}{c^l(t)}\right) = A \left(\frac{d}{c^l(t)}
\right)^{-\alpha} 
\!\!\!\!\!\!= \!
(c^l(t))^{\alpha} 
A 
d^{-\alpha},
\end{equation}
which follows the same power law as $p_0^l$, but scaled by
the factor $(c^l(t))^{\alpha}$.

\begin{remark}
	Note from (\ref{eq:dist-d}) that as $c^l(t)$ increases with $t$, hence $d_i(t)$ increases with $t$.
	However, as we assume that new nodes cannot be added,  the degree cannot grow infinitely
	and (\ref{eq:dist-d}) will not hold for large $t$.
\end{remark}

%%%%%%%%%%%%%%%%%%%%%%%%%%%%%%%%%%%%%%%%

\subsection{Experiments}
\label{sec:expt}

We consider a simple continual learning setting with only two tasks.
Experiments are performed on 
the MNIST data set, with 
the same setup in Section~\ref{sec:mnist-fc}.
The two tasks 
are from
\cite{goodfellow2013empirical,kirkpatrick2017overcoming}.
Task A uses the original images, while 
task B uses
images in which
pixels inside a central $P \times P$ 
square 
are permuted. 
As in \cite{kirkpatrick2017overcoming}),
we use $P=8$ and 26. Note that as the size of MNIST image is $28\times 28$, 
when $P=26$,
the task B
(permuted) 
images 
are very different from the task A 
(original)
images.

\begin{figure}[htbp]
	\begin{center}
		\includegraphics[width=0.485\textwidth]{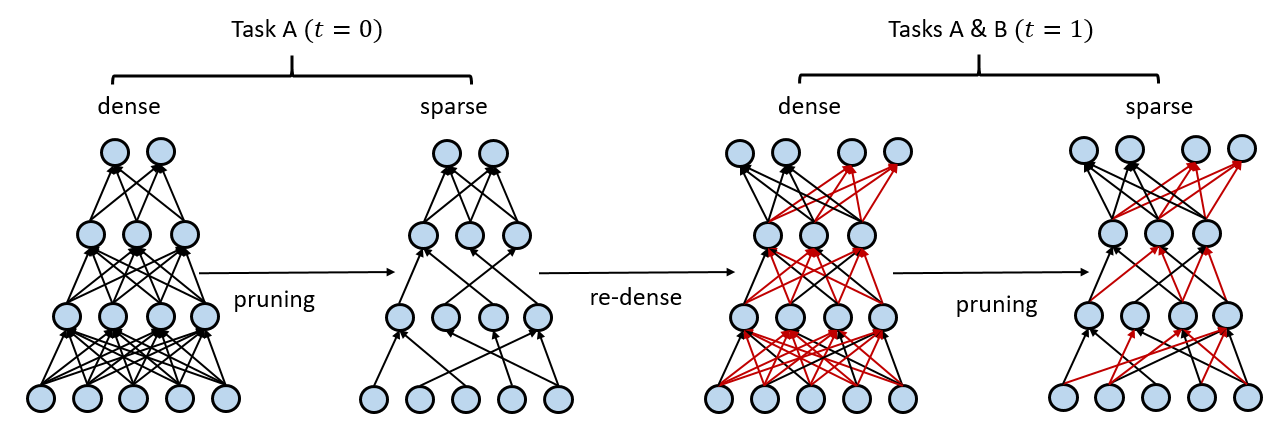}
		\caption{The continual learning setup. 
			Connections learned for task A are
			in black, and those
			for task B
			are in red.
			Left:
			A dense network 
			is trained
			on task
			A; Middle left:  
			Network 
			pruned and 
			retrained on task A;
			Middle right: 
			The pruned connections 
			are reinitialized
			and 
			retrained
			on task B; Right: 
			Connections are
			pruned and the remaining connections
			retrained on task B.
		}
		\label{fig:model_prefer}
	\end{center}
\end{figure}

We first train a dense network on task A
(Figure~\ref{fig:model_prefer}).
A fraction of 
$s_1=0.9$ 
connections 
are pruned
from each layer\footnote{As mentioned at the beginning of Section~\ref{sec:powerlaw},
	connections to the last softmax layer are not pruned.}
as in Section~\ref{sec:powerlaw}.
The remaining connections are
retrained on task A,
and then frozen.
Next, we reinitialize the pruned connections and train a dense network on task B. 
Another fraction of 
$s_2=0.8$
connections
from 
each layer 
are again pruned
and the remaining connections 
retrained.
Recall that connections learned on task A 
are frozen, and so they will not 
be pruned or retrained.
Table~\ref{tbl:fc} shows the testing accuracies of the four networks in Figure~\ref{fig:model_prefer}.  As can be seen,  the sparse network has good performance on both tasks.

\begin{table}[htb]
%	\vspace{-0.2in}
	\centering
	\caption{Testing accuracies (\%) for networks in the continual learning experiment.}
	\label{tbl:fc}
	\begin{tabular}{ c|c|c|c|c} \hline
		& \multicolumn{2}{c|}{task A} & \multicolumn{2}{c}{task B} \\ \cline{2-5}
		$P$ & dense  & sparse  & dense  & sparse  \\ \hline
		8   & 98.09 & 98.21 & 98.17 & 98.09 \\ \hline
		26  & 98.09 & 98.21 & 98.04 & 98.16 \\ \hline
	\end{tabular}
\end{table}

%%%%%%%%%%%%%%%%%%

\subsubsection{Existence of Internal Preferential Attachment}

\begin{figure}[h]
	\begin{center}
		\subfigure[input-to-fc1.
		\label{fig:hidden1024_8_pi_l1}]
		{\includegraphics[width=0.2\textwidth]{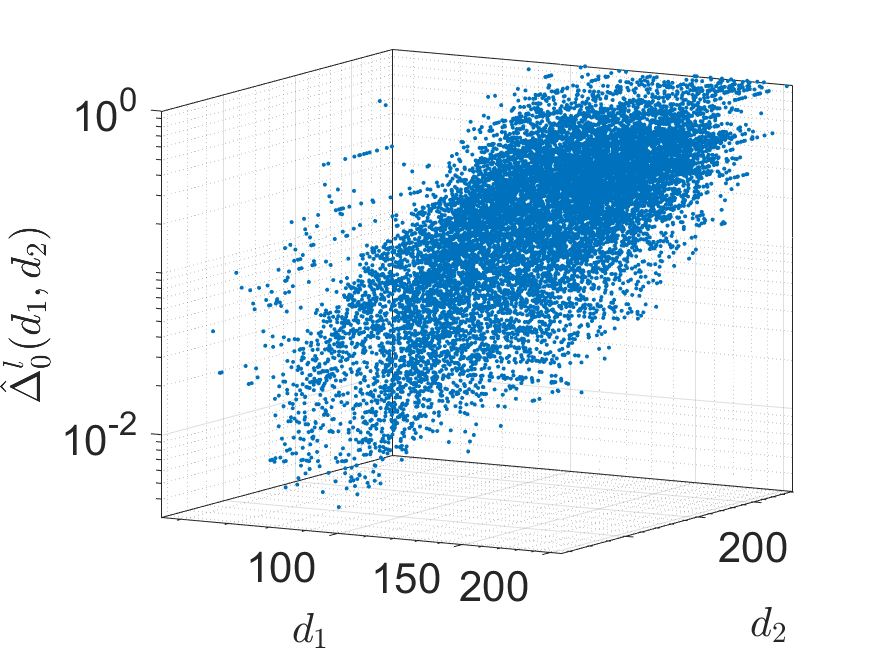}}
		\subfigure[fc1-to-fc2.
		\label{fig:hidden1024_8_pi_l2}]{
			\includegraphics[width=0.2\textwidth]{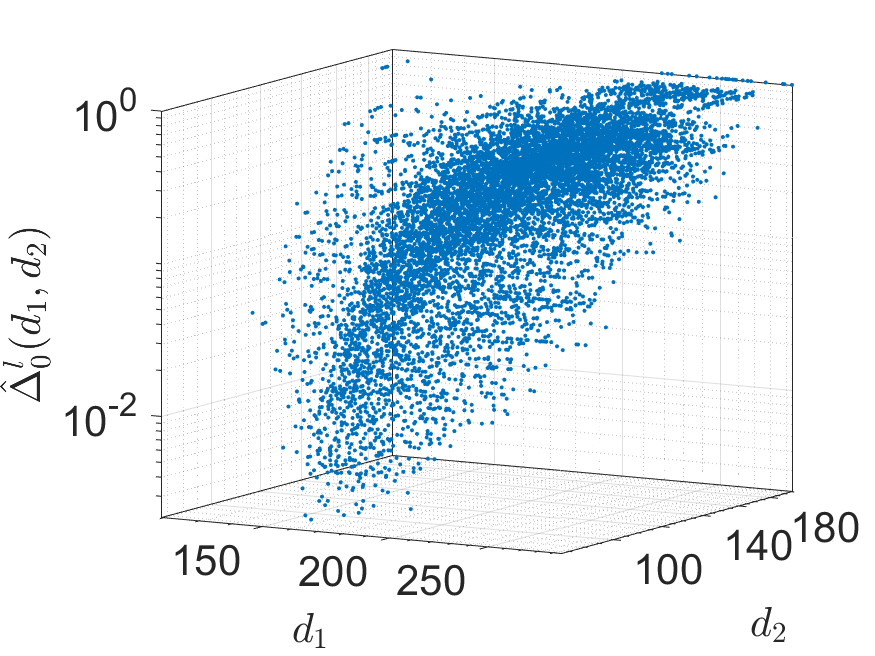}}\\
		\subfigure[input-to-fc1.
		\label{fig:hidden1024_26_pi_l1}]
		{\includegraphics[width=0.2\textwidth]{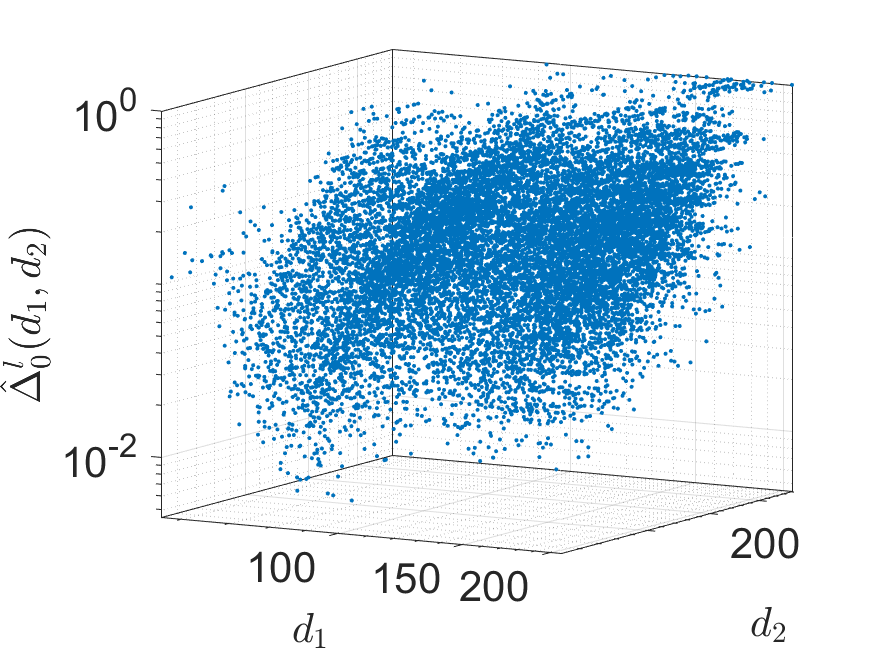}}
		\subfigure[fc1-to-fc2.
		\label{fig:hidden1024_26_pi_l2}]{\includegraphics[width=0.2\textwidth]{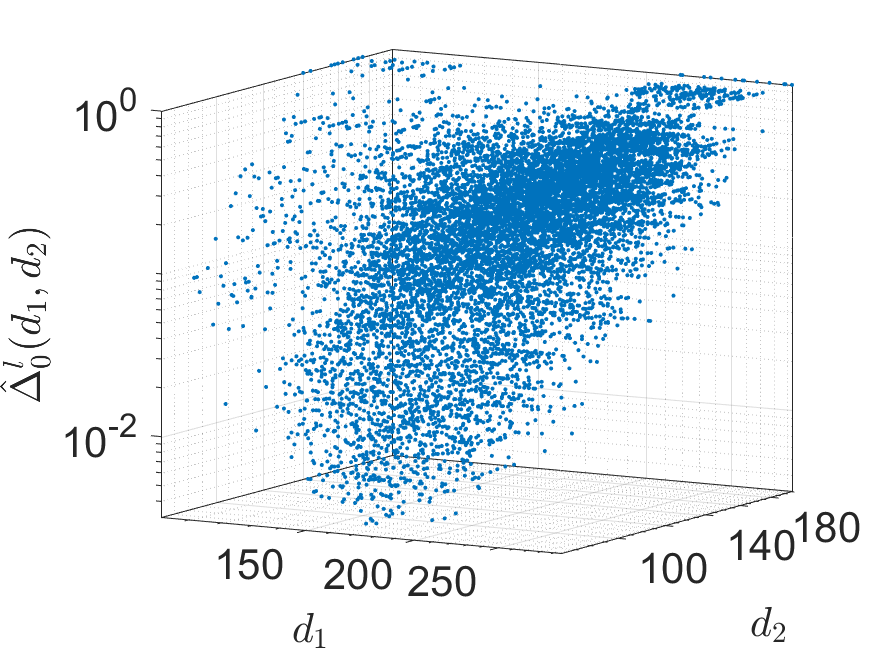}}
		\caption{$\hat{\Delta}_0^l(d_1, d_2)$ vs $d_1$ and $d_2$.
			%Here, ``i" stands for input.
			Top: $P=8$; Bottom: $P=26$.}
		\label{fig:3d}
	\end{center}
\end{figure}

Empirically,
$\Delta_t^l(d_1, d_2)$ 
in (\ref{eq:new})
can be 
estimated by
%\begin{equation}
%\label{eq:delta_new}\hat{\Delta}_t^l(d_1,d_2) = \frac{R_t^l(d_1,d_2)}{ D_t^l(d_1,d_2)},
%\end{equation}
%where \footnote{*** makes intuitive to use $R_{t+1}^l(d_1,d_2)$} $R_t^l(d_1,d_2)$ is 
counting the number of new connections 
created at time $(t+1)$
between 
all the involved node pairs  (i.e., 
one 
from layer $l$ 
with degree $d_1$ 
and the other from
layer $(l+1)$
with degree $d_2$ at time $t$), and then divide it by
%and $D_t^l(d_1,d_2)$ is 
the number of such
node pairs.
%existing at time $t$.
Figure~\ref{fig:3d} shows this empirical estimate  at time $t=0$
(denoted $\hat{\Delta}_0^l(d_1, d_2)$)
versus
the degrees
$d_1$ and $d_2$.
%at $t=0$.
As can be seen,
$\hat{\Delta}_0^l(d_1d_2)$ 
increases with $d_1$ and $d_2$, indicating the presence of
internal preferential attachment.

%Figure~\ref{fig:2d} shows $\hat{\Delta}_0^l(d_1, d_2)$ versus the product $d_1 d_2$. As can be seen, $\hat{\Delta}_0^l(d_1,d_2)$ increases roughly linearly with $d_1d_2$ as assumed in (\ref{eq:new}).

%\begin{figure}[h]
%\begin{center}
%\subfigure[i-to-fc1.
%\label{fig:hidden1024_8_pi_2d_l1}]
%{\includegraphics[width=0.113\textwidth]{figures/diff_pi/scatter_2D/mnist_2_1024_8_0.png}}
%\subfigure[fc1-to-fc2.
%\label{fig:hidden1024_8_pi_2df_l2}]{
%			\includegraphics[width=0.113\textwidth]{figures/diff_pi/scatter_2D/mnist_2_1024_8_1.png}}
%\subfigure[i-to-fc1.
%\label{fig:hidden1024_26_pi_2d_l1}]
%{\includegraphics[width=0.113\textwidth]{figures/diff_pi/scatter_2D/mnist_2_1024_26_0.png}}
%\subfigure[fc1-to-fc2.
%\label{fig:hidden1024_26_pi_2d_l2}]{
%			\includegraphics[width=0.113\textwidth]{figures/diff_pi/scatter_2D/mnist_2_1024_26_1.png}}
%\caption{$\hat{\Delta}_0^l(d_1, d_2)$ vs $d_1 d_2$.
%Left: $P=8$; Right: $P=26$.}
%\label{fig:2d}
%	\end{center}
%\end{figure}

%%%%%%%%%%%%%%%%%%

\subsubsection{Number of New Connections versus Degree}

From (\ref{eq:tmp}), for a node with degree $d_i(0)$ at 
$t=0$,
the number of connections  added to it at $t=1$ is equal to 
\begin{equation} \label{eq:tmp1}
\left.\frac{\mathrm d d_i(t)}{\mathrm d t}\right|_{t=0} = d_i(0) \frac{N^la^l+N^{l-1}a^{l-1}}{\sum_s d_s(0)}.
\end{equation} 
Let $d_i(0)=d$, 
an empirical estimate of
$\left.\frac{\mathrm d d_i(t)}{\mathrm d t}\right|_{t=0}$ (denoted
$\hat{\Omega}_0^l(d)$)
can be obtained by 
counting the number of new connections that all nodes (from layer $l$) with degree $d$  
made at time $(t+1)$, and then divide it by the number of degree-$d$ nodes (in layer $l$)
at time $t$.

\begin{figure}[htbp]
	\begin{center}
		\subfigure[ input. \label{fig:hidden1024_8_0}]{\includegraphics[width=0.155\textwidth]{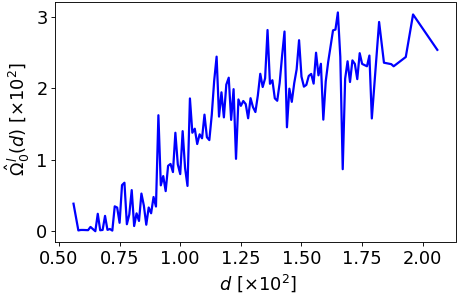}}
		\subfigure[ fc1. \label{fig:hidden1024_8_1}]{\includegraphics[width=0.155\textwidth]{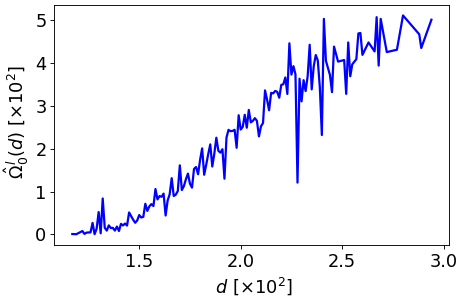}}
		\subfigure[ fc2. \label{fig:hidden1024_8_2}]{\includegraphics[width=0.155\textwidth]{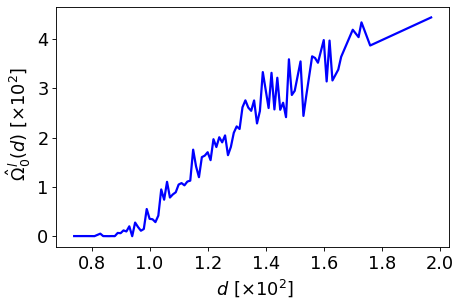}}
		
		\subfigure[ input. \label{fig:hidden1024_26_0}]{\includegraphics[width=0.155\textwidth]{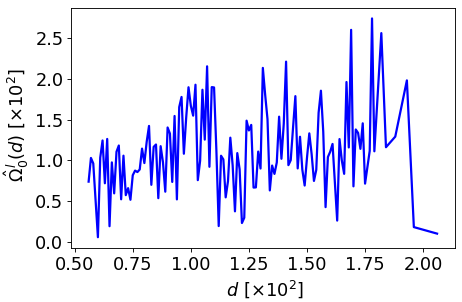}}
		\subfigure[ fc1. \label{fig:hidden1024_26_1}]{\includegraphics[width=0.155\textwidth]{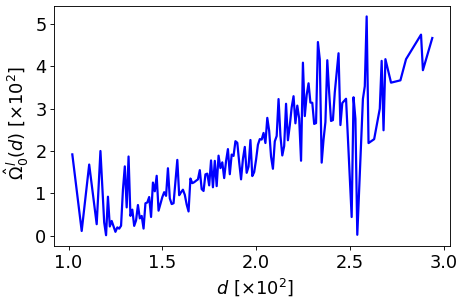}}
		\subfigure[ fc2. \label{fig:hidden1024_26_2}]{\includegraphics[width=0.155\textwidth]{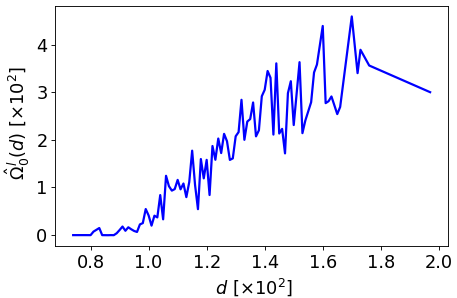}}
%		\vspace{-0.1in}
		\caption{Number of new connections $\hat{\Omega}_0^l(d)$ vs degree $d$.
			Top: $P=8$; Bottom: $P=26$.} 
		\label{fig:preferential_fc}
	\end{center}
%	\vspace{-0.1in}
\end{figure}

Figure~\ref{fig:preferential_fc} shows 
$\hat{\Omega}_0^l(d)$
versus the node degree $d$ at time $t=0$.
As can be seen, 
when the two tasks are similar ($P=8$),
the relationship 
is roughly linear for all layers, which agrees with (\ref{eq:tmp1}).
However,
when the tasks are much less similar ($P=26$), 
the network must 
learn to associate new collections from pixels to penstrokes, and 
connections from the input layer to the first hidden layer have to be
significantly modified.
Hence, preferential attachment is no longer useful.
As can be seen, there is no
linear relationship
for the input layer,
and the 
linear relationship 
in the fc1 layer\footnote{Recall that the 
	fc1 layer also counts the connections to the input layer.}
%and the fc2 layer.}
is noisier 
than that for $P=8$.
On the other hand,
as discussed in \cite{goodfellow2013empirical},
once the input layer has established new associations to map from pixels to
penstrokes,
%because the two tasks are highly similar to each other in terms of the underlying structure with only input presented in a different format.
the higher-level feature extractors (i.e., fc2) are only dependent on these
penstroke features (extracted at fc1) and thus
less affected by the input permutation.
As can be seen,
the fc2 layer still shows a linear relationship.

%By permutation, the original task and the new task benefit from having concepts like penstroke detectors~\cite{goodfellow2013empirical}.  

%%%%%%%%%%%%%%%%%%

\subsubsection{Degree Distribution}

\begin{figure}[htb]
	\begin{center}
		\subfigure[input.
		%({\sf K}=1024). 
		\label{fig:mnist_1024_1_8}]{\includegraphics[width=0.155\textwidth]{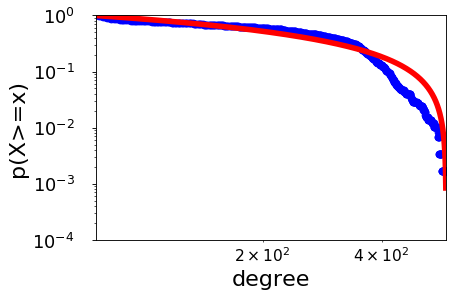}}
		\subfigure[fc1.
		%({\sf K}=1024).
		\label{fig:mnist_1024_2_8}]{\includegraphics[width=0.155\textwidth]{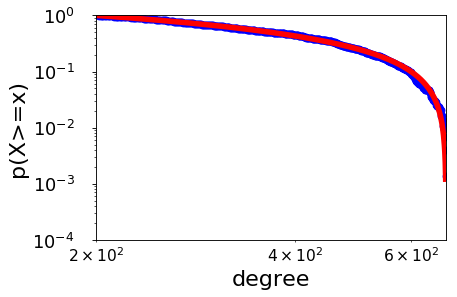}}
		\subfigure[fc2.
		%({\sf K}=1024).
		\label{fig:mnist_1024_3_8}]{\includegraphics[width=0.155\textwidth]{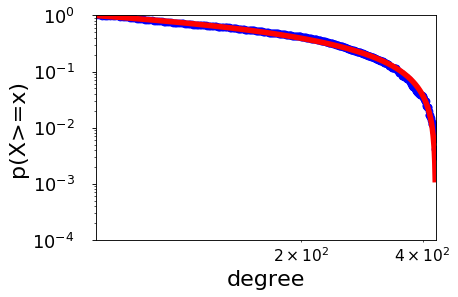}}		
		\subfigure[input.
		%({\sf K}=1024). 
		\label{fig:mnist_1024_1_26}]{\includegraphics[width=0.155\textwidth]{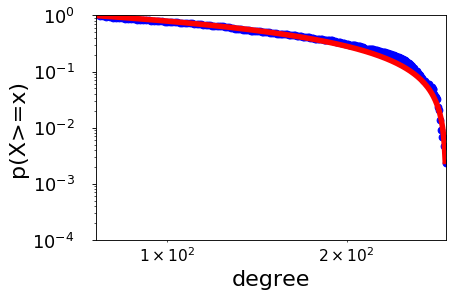}}
		\subfigure[fc1.
		%({\sf K}=1024).
		\label{fig:mnist_1024_2_26}]{\includegraphics[width=0.155\textwidth]{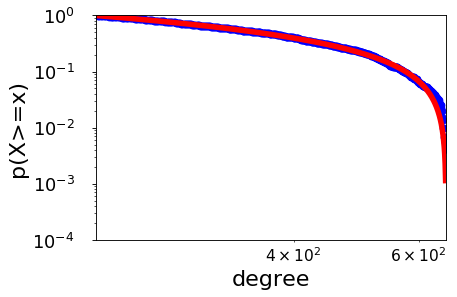}}
		\subfigure[fc2.
		%({\sf K}=1024).
		\label{fig:mnist_1024_3_26}]{\includegraphics[width=0.155\textwidth]{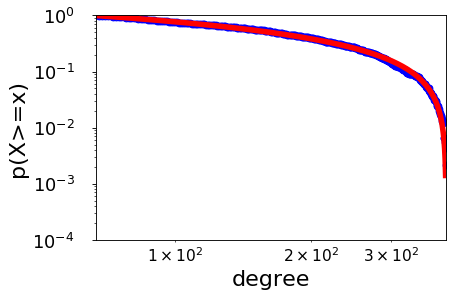}}		
		\caption{ Log-log CCDF plot and TPL fit (red) for the MLP layers on MNIST at $t=1$. Top: $P=8$; Bottom: $P=26$.
			%			Top: {\sf K}=400; middle: {\sf K}=1024; bottom: {\sf K}=2048.
			%			The fitted exponent is shown at the top of each subplot.   
		}
%		\vspace{-0.1in}
		\label{fig:fc_ccdf_new}
	\end{center}
%	\vspace{-0.1in}
\end{figure}

Recall 
from Section~\ref{sec:mnist-fc}
that 
the degree distribution 
of the sparse MLP trained on task A ($t=0$)
follows the TPL.
From (\ref{eq:degree-t}),
%if the initial $(t=0)$ degree distribution follows the power law,
we thus expect that its degree distribution 
after training on task B ($t=1$)
also follows the TPL.
Figure~\ref{fig:fc_ccdf_new} 
%and Table~\ref{tbl:mnist_mlp_8_26_fitting} 
shows the CCDF plot
and TPL fit for each
layer of the sparse network at $t=1$.
%, respectively.
%, and Table~\ref{tbl:mnist_mlp_8_26_fitting} shows parameters of the TPL fits.
	As can be seen, the $p$-values of the first layer are relatively low.
	This is because that, due to permutation, the input layer has to establish new associations to map from pixels to penstrokes.	
%	Moreover, as\footnote{*** so what's the consequence?} high degree nodes tend to have a lower\footnote{*** relatively
%	lower only if the max deg has been reached. is that really the case?} probability of making connections due to the limited capacity. 
%Another reason is that degree-$0$ nodes at $t=0$ also have an opportunity to form
%connections at $t=1$, while the model in Section~\ref{sec:pref} assumes that
%preferential attachment only over nodes with degree larger than or equal to
%1.

%\begin{table}[htbp]
%	\centering
%	\caption{Power-law fits for different MLP layers (on MNIST data) for $t=1$  and
%	their $p$-values.}
%	\label{tbl:mnist_mlp_8_26_fitting}
%	\begin{tabular}{ccc|ccc|c}
%		\hline
%		$P$ & layer & $n$  & $x_{\min}$ & $x_{\max}$ & $\alpha$ & $p$  \\ \hline
%		    & \textit{input} & \textit{784} &     \textit{76}      &      \textit{582}     &  \textit{1.001}  & \textit{0.00} \\
%		 8  &  fc1  & 1024 &       201        &       677        &  1.008   & 0.76 \\
%		    &  fc2  & 1024 &        62        &       431        &  1.007   & 0.24 \\ \hline
%		    & input & 784  &        76        &       294        &  1.001   & 0.20 \\
%		26  &  fc1  & 1024 &       211        &       656        &  1.280   & 0.99 \\
%		    &  fc2  & 1024 &        67        &       397        &  1.003   & 0.52 \\ \hline
%	\end{tabular}
%\end{table}

%%%%%%%%%%%%%%%%%%%%%%%%%%%%%%%%%%%%%%%%%%%%%%%%%%%%%%%%%%%%%%%%%%%%%%

\section{Conclusion}
\label{sec:conc}

In this paper, we showed that 
a number of sparse deep learning models exhibit the 
(truncated) power law behavior.
We also proposed an internal preferential attachment model to explain how the network topology
evolves,
and verify in a continual learning setting that new connections added
to the network indeed have this preferential bias.

In biological neural networks
with limited capacities, 
reuse of neural circuits is essential for learning multiple tasks,
and scale-free networks can
%resemble them most and meanwhile 
learn faster than random and small-world networks
\cite{monteiro2016model}.
%for various cognitive purposes is recognized as a central organizational principle, and scale-free networks following preferential attachment are shown to exhibit fastest learning compared with other network topology. 
In the future, 
we will use  the 
dynamic behavior
observed 
on artificial neural networks
to design faster continual learning algorithms.
%which is desirable in artificial general intelligence, 
%exaptation of  the learned connectivity (e.g. evolving the network following preferential attachment).
Moreover, this paper only studies feedforward neural networks. We will also study
existence 
of the power law 
in recurrent neural networks.
%where connections share across time, instead of space as in %convolutional networks.  

%Our findings help understanding the connectivity patterns in deep neural networks.
\bibliography{paper}
\end{document}